\documentclass[conference]{IEEEtran}
\usepackage{times}

\usepackage[numbers]{natbib}
\usepackage{multicol}
\usepackage[bookmarks=true]{hyperref}
\usepackage[compatibility=false]{caption}
\usepackage{acronym}
\usepackage{xspace}
\usepackage{graphicx}
\usepackage{caption}
\usepackage{amsmath}
\usepackage{array}
\usepackage{xcolor}
\usepackage{subcaption}
\usepackage{booktabs}
\usepackage{comment}
\usepackage{tabularx}
\usepackage{multirow}
\usepackage{float}
\usepackage{dsfont}
\usepackage{siunitx}
\let\labelindent\relax
\usepackage{enumitem}
\usepackage{wrapfig}
\definecolor{LightGray}{gray}{0.9}
\definecolor{cvprblue}{rgb}{0.21,0.49,0.74}
\usepackage[capitalise,nameinlink]{cleveref}

\makeatletter
\renewcommand{\paragraph}{%
  \@startsection{paragraph}{4}%
  {\z@}{0ex \@plus 0ex \@minus 0ex}{-1em}%
  {\hskip\parindent\normalfont\normalsize\bfseries}%
}
\makeatother

\acrodef{ddim}[DDIM]{Denoising Diffusion Implicit Models}
\acrodef{mlp}[MLP]{multi-layer perception}
\acrodef{bc}[BC]{Behavior cloning}
\acrodef{mpc}[MPC]{model predictive control}
\acrodef{rl}[RL]{reinforcement learning}
\acrodef{il}[IL]{imitation learning}
\acrodef{wbc}[WBC]{Whole body control}
\acrodef{llm}[LLM]{large language model}
\acrodef{mmlm}[MMLM]{multi-modal language model}
\acrodef{sft}[SFT]{supervised fine-tuning}
\acrodef{rlhf}[RLHF]{Reinforcement Learning from Human Feedback}
\acrodef{sg3d}[3DSG]{3D Scene Graph}
\acrodef{qa}[QA]{question-answering}
\acrodef{gui}[GUI]{graphical user interface}
\acrodef{cot}[CoT]{Chain-of-Thought}
\acrodef{mse}[MSE]{mean squared error}
\acrodef{ppo}[PPO]{Proximal Policy Optimization}

\newcommand{\modelname}{\textsc{OmniTrack}\xspace}

\definecolor{gblue}{HTML}{4285F4}
\definecolor{gred}{HTML}{DB4437}
\definecolor{ggreen}{HTML}{0F9D58}
\definecolor{vblue}{HTML}{2993ba}

\newcommand{\goalstate}{{\bs{o}^{\text{g}}_t}}

\newcommand{\selfstate}{{\bs{o}^{\text{p}}_t}}

\newcommand{\state}{{\bs{o}_t}}
\newcommand{\action}{{\bs{a}_t}}

\newcommand{\simp}{{\bs{{q}}_{t}}}
\newcommand{\simv}{{\bs{\dot{q}}_{t}}}

\newcommand{\simav}{{\bs{{\omega}}_{t}}}

\newcommand{\bs}[1]{\boldsymbol{#1}}

\crefname{algorithm}{Alg.}{Algs.}
\Crefname{algocf}{Algorithm}{Algorithms}
\crefname{section}{Sec.}{Secs.}
\Crefname{section}{Section}{Sections}
\crefname{table}{Tab.}{Tabs.}
\Crefname{table}{Table}{Tables}
\crefname{figure}{Fig.}{Fig.}
\Crefname{figure}{Figure}{Figure}
\crefname{equation}{Eq.}{Eq.}
\Crefname{equation}{Equation}{Equation}

\usepackage{amsmath,amsfonts,bm}

\def\eqref#1{equation~\ref{#1}}

\def\1{\bm{1}}

\DeclareMathAlphabet{\mathsfit}{\encodingdefault}{\sfdefault}{m}{sl}
\SetMathAlphabet{\mathsfit}{bold}{\encodingdefault}{\sfdefault}{bx}{n}

\begin{document}

\title{OmniTrack: General Motion Tracking \\ via Physics-Consistent Reference}
\author{
\renewcommand{\arraystretch}{1.08}
\begin{tabular}{ccc}
Yuhan Li\textsuperscript{1,2}$^{*}$ &
Peiyuan Zhi\textsuperscript{2}$^{*}$ &
Yunshen Wang\textsuperscript{2,3} \\
Tengyu Liu\textsuperscript{2} &
Sixu Yan\textsuperscript{1,2} &
Wenyu Liu\textsuperscript{1} \\
Xinggang Wang\textsuperscript{1}$^{\dagger}$ &
Baoxiong Jia\textsuperscript{2}$^{\dagger}$ &
Siyuan Huang\textsuperscript{2}$^{\dagger}$ \\
\end{tabular}
\vspace{5pt}\\
{\small
\renewcommand{\arraystretch}{1.05}
\begin{tabular}{c}
\textsuperscript{1} Huazhong University of Science and Technology \\
\textsuperscript{2} State Key Lab of General AI, Beijing Institute for General Artificial Intelligence (BIGAI) \\
\textsuperscript{3} Shanghai Jiao Tong University
\end{tabular}
}
\vspace{4pt}\\
{\small
$^{*}$ Equal contribution \quad
$^{\dagger}$ Corresponding authors \quad 
Project Website: \url{https://omnitrack-humanoid.github.io/}
}
\vspace{-10pt}
}

\twocolumn[{
\renewcommand\twocolumn[1][]{#1}
\maketitle
\begin{center}
  \vspace{1em}
  \captionsetup{type=figure}
  \includegraphics[width=\linewidth]{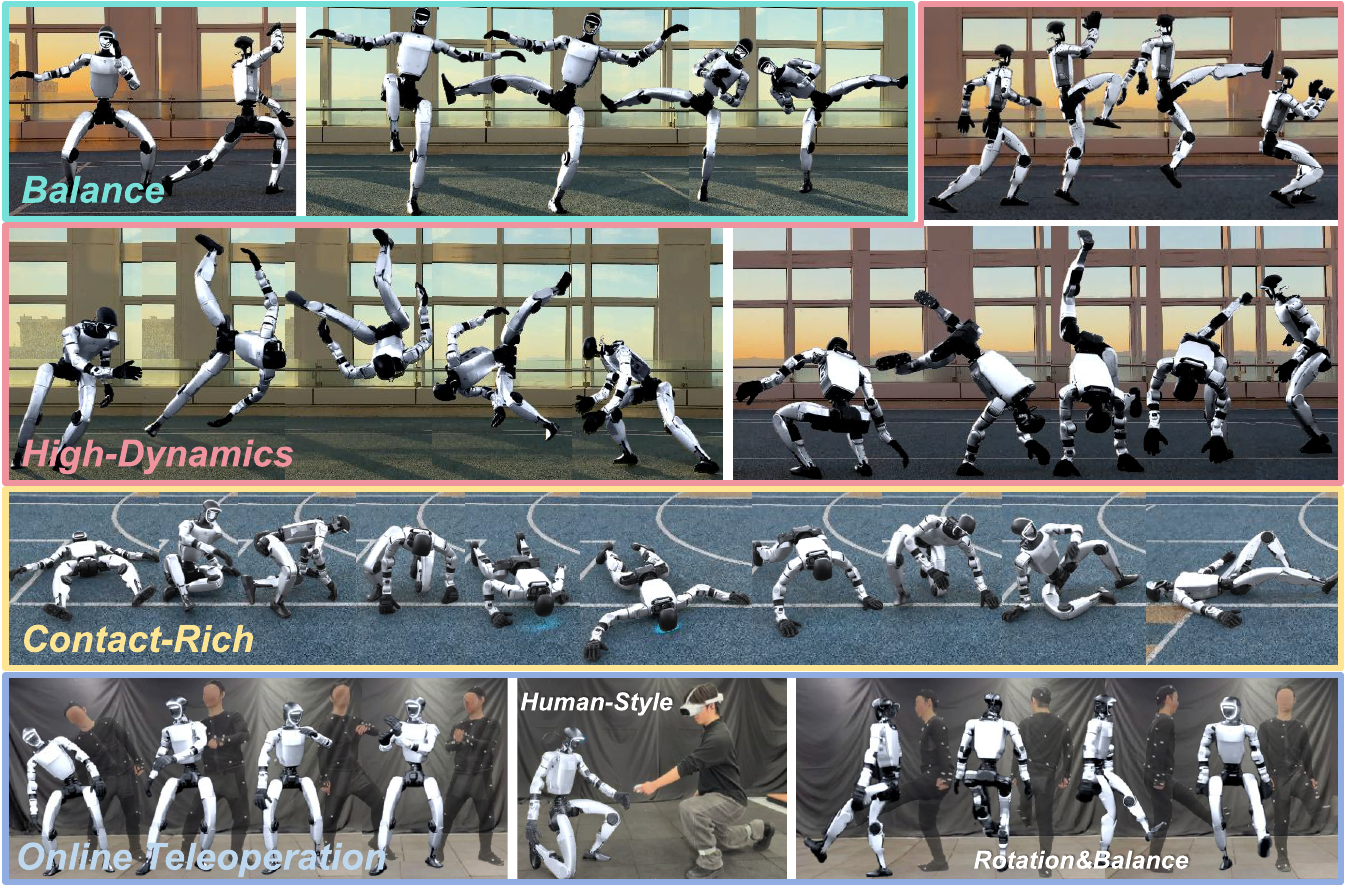}
  \captionof{figure}{
    \textbf{Capabilities of \modelname in general motion tracking and real-time teleoperation.} Leveraging physics-consistent reference motions, OmniTrack achieves general tracking across diverse motion categories, including balance control, high-dynamic maneuvers, and contact-rich interactions. OmniTrack supports real-time teleoperation, enabling the execution of diverse human-style dynamic movements as well as interactive behaviors. These results demonstrate the generality of the framework in handling both physically demanding motions and unstructured motion commands.
  }
  \label{fig:teaser}
\end{center}
}]

\begin{abstract}

Learning motion tracking from rich human motion data is a foundational task for achieving general control in humanoid robots, enabling them to perform diverse behaviors. However, discrepancies in morphology and dynamics between humans and robots, combined with data noise, introduce physically infeasible artifacts in reference motions, such as floating and penetration. During both training and execution, these artifacts create a conflict between following inaccurate reference motions and maintaining the robot's stability, hindering the development of a generalizable motion tracking policy.
To address these challenges, we introduce \modelname, a general tracking framework that explicitly decouples physical feasibility from general motion tracking. In the first stage, a privileged generalist policy generates physically plausible motions that strictly adhere to the robot's dynamics via trajectory rollout in simulation. In the second stage, the general control policy is trained to track these physically feasible motions, ensuring stable and coherent control transfer to the real robot. 
Experiments show that \modelname improves tracking accuracy and demonstrates strong generalization to unseen motions. In real-world tests, \modelname achieves hour-long, consistent, and stable tracking, including complex acrobatic motions such as flips and cartwheels. Additionally, we show that \modelname supports human-style stable and dynamic online teleoperation, highlighting its robustness and adaptability to varying user inputs.

\end{abstract}

\IEEEpeerreviewmaketitle

\section{Introduction}
Learning general humanoid control from human motion data has emerged as a dominant paradigm for enabling diverse and human-like behaviors \cite{ji2024exbody2, he2024omnih2o, luo2025sonic,Kalaria2025DreamControlHW, mnih2015human,li2025language,margolis2025softmimic}. By tracking reference motions collected from large-scale human demonstrations, recent methods have shown promising results across a variety of skills \cite{he2025asap, xie2025kungfubot, liao2025beyondmimic, chen2025gmt, li2025bfm}. However, these approaches are still largely limited to short-term motion sequences\cite{he2025asap, xie2025kungfubot, chen2025gmt}. In particular, existing policies have difficulty generalizing across heterogeneous motion types, often facing inherent trade-offs between stable behaviors and highly dynamic actions, as well as between everyday movements and contact-rich interactions.

\begin{figure}[!t]
    \centering
    \includegraphics[width=\columnwidth]{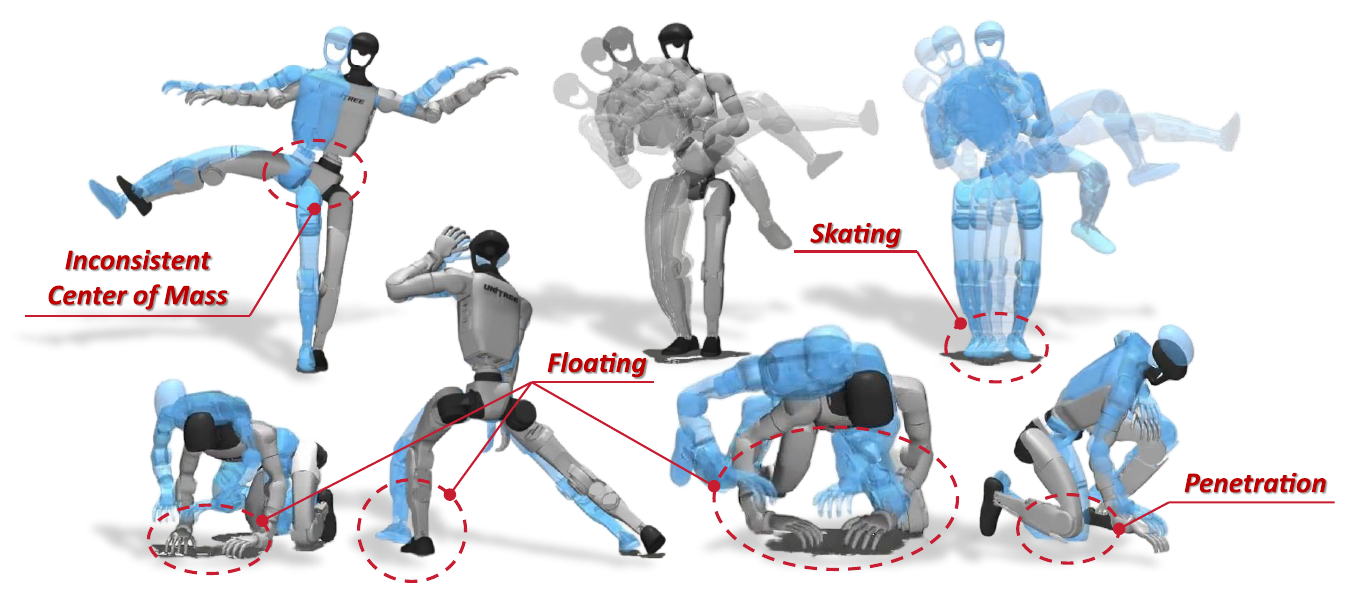}
    \caption{Examples of physically infeasible artifacts in retargeted human motions (blue) compared with physically feasible robot motions (black/gray), including inconsistent center-of-mass motion, foot skating, floating, and ground penetration, which hinder stable humanoid control.}
    \label{fig:phsical artifacts}
\end{figure}

A key limitation of prior works stems from the fundamental embodiment gap \cite{yang2025omniretarget} between human demonstrators and humanoid robots. Differences in kinematics, body proportions, and degrees of freedom, further compounded by variations in motion capture hardware and preprocessing pipelines, introduce physically infeasible artifacts in reference motions, such as floating and penetration, as shown in \cref{fig:phsical artifacts}. Some existing works~\cite{lee2025phuma,yang2025omniretarget,pan2025spiderscalablephysicsinformeddexterous} attempt to mitigate these physical artifacts through motion retargeting. However, the retargeting process typically considers only kinematic constraints and does not explicitly model dynamic consistency, and therefore cannot fundamentally eliminate the \textbf{physical inconsistencies present in the reference motions}. The difficulty of policy learning is coupled with the magnitude of these artifacts, creating an intrinsic conflict between high-fidelity tracking and physical stability. The optimal trade-off between these objectives varies unpredictably across motions, implicitly requiring the policy to discern physical feasibility. However, constrained by partial observability in real-world settings, the robot lacks the privileged information required to make such judgments, such as global position and velocity. As a result, it blindly pursues tracking rewards even when references are infeasible (e.g., floating, penetration), leading to frequent dynamic failures and ultimately preventing the convergence of a truly generalist control policy.


To overcome the limitation, we propose \modelname, a general motion tracking framework that \textit{explicitly decouples physical feasibility from motion tracking}. This design enables a single policy to perform diverse and complex behaviors—such as sprinting, side cartwheels, and dancing—while remaining stable over long-horizon motions, as shown in \cref{fig:teaser}. \modelname is built on the insight that, under partial observability, determining the physical feasibility of reference motions is inherently difficult. Rather than requiring a single controller to implicitly balance tracking fidelity and physical stability, we address physical feasibility before policy learning. By enforcing physical consistency in advance, the controller is freed from compensating for infeasible artifacts during execution. Following this principle, \modelname adopts a two-stage training pipeline that removes the influence of physical artifacts, enabling stable and generalizable humanoid motion control.

The physical motion generation stage \textbf{converts raw reference motions into physically feasible motions}. We train a generalist policy with full privileged information and use it solely as a motion generation module. With access to complete simulation states, the policy can accurately assess physical feasibility and reconstruct reference motions through trajectory rollouts in a physics simulator, producing state trajectories that strictly comply with the robot’s dynamics. Since this stage aims to obtain high-precision, dynamics-consistent reference trajectories, we exclude observation noise and domain randomization during training to maximize reconstruction fidelity. Moreover, this process provides high-quality supervisory signals that are difficult to obtain from raw motion capture data, such as accurate per-link contact masks, which are crucial for the stable convergence of the subsequent general control policy. More importantly, the \textbf{physically plausible motion forms a standardized benchmark}. With physical artifacts removed, performance differences more directly reflect true tracking capability rather than the ability to compensate for flawed references, enabling fairer and more interpretable evaluation.

The general motion tracking stage is designed to \textbf{achieve robust tracking of diverse behaviors on physical hardware}. We train a single general control policy using the physically feasible motions generated in the previous stage. To reflect real-world deployment conditions, the policy operates under partial observability, hardware constraints, and unmodeled dynamics, relying only on onboard sensors. Because the reference motions are already dynamics-consistent, the intrinsic trade-off between tracking fidelity and physical stability is largely alleviated. This allows the policy to focus on learning generalist motion tracking and sim-to-real transfer, rather than compensating for physically infeasible artifacts present in the original reference motions.

Systematic experiments on the Unitree-retargeted LAFAN1~\cite{harvey2020robust} and AMASS\cite{mahmood2019amass} datasets demonstrate that \modelname significantly outperforms existing baselines in motion-tracking accuracy, with particularly pronounced advantages in highly dynamic and contact-rich scenarios. Our method also exhibits strong generalization to unseen motion sequences. More importantly, we validate the real-world effectiveness of \modelname on the Unitree G1 humanoid robot, \textbf{achieving zero-shot sim-to-real transfer across the entire LAFAN1 motion repertoire}. The policy maintains \textbf{hour-long, continuous, and stable tracking}, and demonstrates high dynamic capability by reliably executing challenging acrobatic motions such as \textbf{flips and cartwheels}. In addition, \modelname supports \textbf{human-style stable and dynamic online teleoperation}, showing robust responsiveness to continuous, user-driven inputs and the ability to handle unpredictable online commands.

Our main contributions are summarized as follows:

\begin{enumerate}[leftmargin=*,noitemsep,nolistsep,topsep=0pt,partopsep=0pt]
    \item 
    
    We propose a two-stage learning framework that explicitly decouples physical feasibility from general motion tracking, enabling a single policy to continuously execute hour-long diverse motions.
    
    \item 
    We construct a high-quality, physically consistent robot motion dataset by removing dynamic artifacts from raw human motion data, providing a standardized benchmark for general humanoid control.

    \item 
    
    We demonstrate robust zero-shot sim-to-real transfer on a humanoid robot, including strong generalization under human-style, dynamic real-time teleoperation and reliable execution of highly dynamic acrobatic behaviors.
\end{enumerate}

\section{Related Work}

\subsection{Motion Retargeting for Humanoids}
Human motion data has been widely adopted in both physics-based character control and humanoid control, where it is commonly retargeted to drive virtual characters \cite{peng2018deepmimic, peng2021amp, luo2023perpetual, tessler2024maskedmimic,feng2025physhmr} or real humanoid systems\cite{he2024omnih2o, li2025clone, liao2025beyondmimic, he2025asap, chen2025gmt, allshire2025visual,weng2025hdmi,zeng2025behavior,liu2025trajbooster,sun2025ulc,hu2025slac,wang2025experts,kuang2025skillblender,li2025learning,li2025amo,shi2025adversarial,ben2025homie}. Keypoint-based or orientation-based retargeting methods \cite{luo2023perpetual,araujo2025retargeting} help align human motions with target embodiments, but the inherent morphological gap between humans and humanoid robots often leads to physically infeasible artifacts,, such as floating, foot skating, penetration, and imbalance. A central challenge, therefore, lies in ensuring that the retargeted motions remain physically plausible and executable. Recent works have explored physics-aware retargeting to address this issue. PHUMA~\cite{lee2025phuma} enforces physical feasibility through careful data curation and physics-constrained optimization, explicitly handling joint limits, ground contact consistency, and foot-slip suppression to obtain large-scale motion datasets that remain physically reliable. OmniRetarget~\cite{yang2025omniretarget} adopts a deformation-based formulation that minimizes Laplacian distortion between human and robot meshes under kinematic constraints, producing trajectories that are kinematically feasible for humanoids. While effective in reducing geometric inconsistencies, optimization-based pipelines still cannot fully guarantee physical realizability during execution, and even minor residual artifacts can significantly hinder downstream policy learning. In contrast, \modelname adopts a reinforcement-learning-based approach that generates physically plausible motions directly through simulated rollouts. By training policies within a physics simulator and treating simulated states as motion data, \modelname enforces physical feasibility through interaction with the dynamics rather than post-hoc optimization. This produces smooth, dynamically consistent trajectories with minimal jitter, offering a more reliable foundation for subsequent control or imitation learning.

\subsection{General Motion Tracking for Humanoid}
In general motion tracking, recent methods have demonstrated rich human-like behaviors~\cite{cheng2024expressive,ji2024exbody2,he2025asap,chen2025gmt,zhang2025track,yin2025unitracker,liao2025beyondmimic,yue2025rl,han2025kungfubot2,zhao2025resmimic,wang2025physhsi} and increasingly general tracking capabilities~\cite{he2024omnih2o,luo2025sonic}, enabling challenging motion reproduction and downstream teleoperation tasks~\cite{li2025clone,ze2025twist,ze2025twist2}. Despite this progress, achieving robust and general motion tracking on real humanoid hardware remains difficult due to embodiment mismatch, actuator limits, and sim-to-real discrepancies. Existing approaches reflect different trade-offs. ExBody2~\cite{ji2024exbody2} and OmniH2O~\cite{he2024omnih2o} mainly track quasi-static behaviors; ASAP~\cite{he2025asap} handles dynamic motions but only over short horizons. GMT~\cite{chen2025gmt} improves generality but struggles with contact-rich sequences, while UniTracker~\cite{yin2025unitracker} relies on dedicated adaptation modules. BeyondMimic~\cite{liao2025beyondmimic} excels on single clips but lacks long-horizon robustness. AnyTrack~\cite{zhang2025track} introduces an adaptation module but remains sensitive to low-quality motion inputs. Sonic~\cite{luo2025sonic} achieves strong tracking performance but demands large-scale datasets and substantial computational resources.

A particularly critical limitation shared by many approaches is their sensitivity to low-quality or noisy reference motions. In online settings such as teleoperation, artifacts such as jitter, inconsistent contacts, and unstable joint transitions can directly propagate to the controller, leading to instability and safety risks. By explicitly decoupling physical feasibility from general motion tracking, our two-stage framework significantly improves long-horizon stability and physical plausibility, enabling robust humanoid motion tracking even under degraded inputs and challenging real-world conditions.

\section{Method}

\begin{figure*}[!t]
    \centering
    \includegraphics[width=\textwidth]{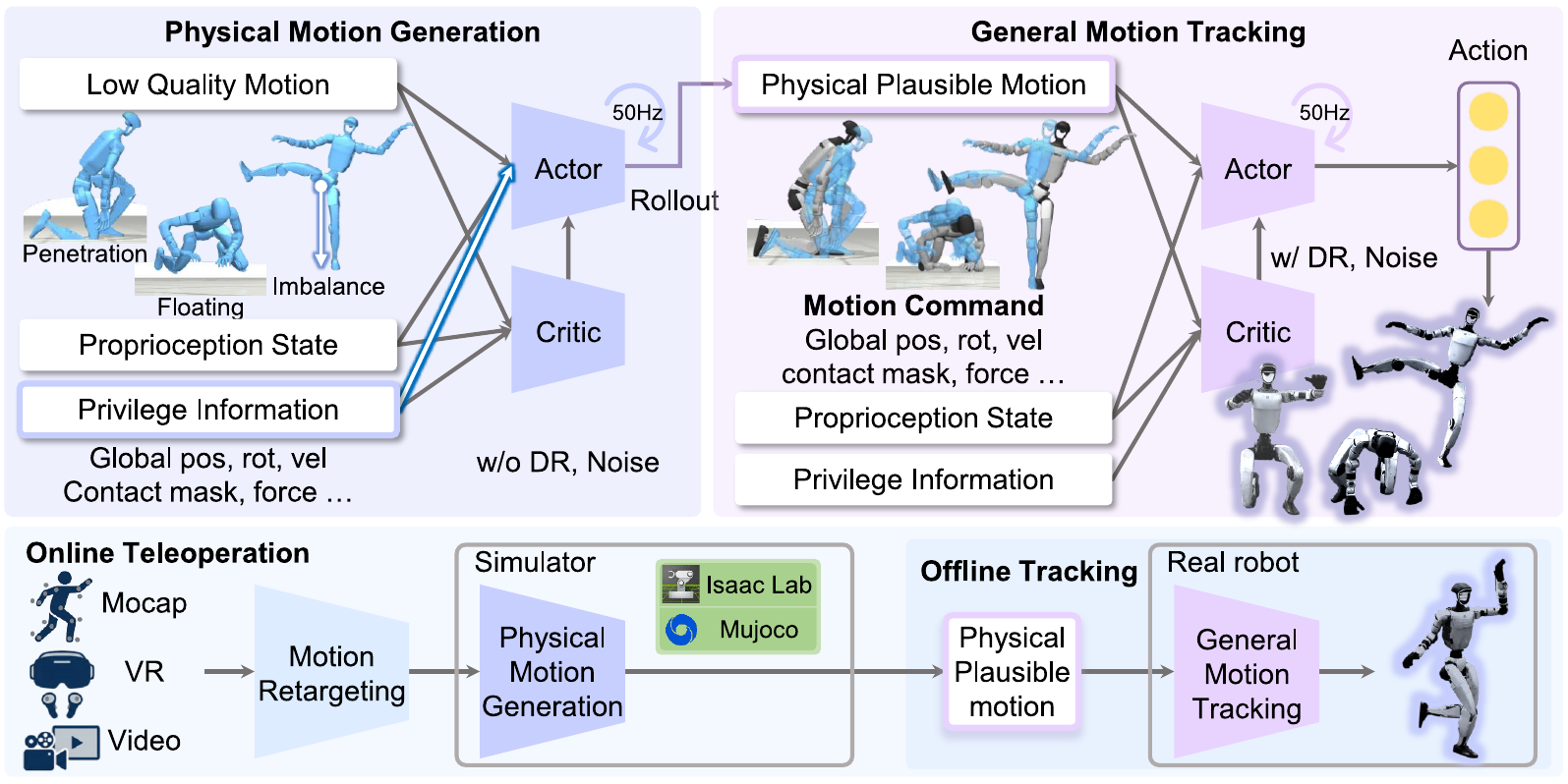}
    \caption{\textbf{Overview of the \modelname framework.} Our method adopts a two-stage pipeline that first converts raw, physically inconsistent reference motions into physics-consistent motions in simulation, and then trains a general policy to robustly track these motions under realistic conditions. The resulting system supports both offline motion tracking and online teleoperation.}
    \label{fig:framework}
\end{figure*}

We propose \modelname, a two-stage framework for learning a general humanoid motion tracking policy that explicitly decouples \emph{physical feasibility} from \emph{general motion tracking}, as illustrated in~\cref{fig:framework}. The core idea is to first eliminate physically infeasible artifacts caused by the embodiment gap at the level of motion references, and then learn a general control policy solely on physics-consistent trajectories under realistic sensing conditions.

\noindent\textbf{Problem Formulation}
We formulate the learning in both stages as Markov Decision Processes and optimized using PPO \cite{schulman2017proximal}. At each timestep, the policy receives the current robot state $\state$ and a motion command $\goalstate$ derived from a reference trajectory, and outputs target joint positions $\action$, which are tracked by joint-level PD controllers. We adpot the common tracking reward formulation  $r_t=r_{\text{track}} + r_{\text{reg}}$, where $r_{\text{track}}$ penalizes tracking in root and link poses, orientations, and velocities with respect to the reference motion, while $r_{\text{reg}}$ regularizes learning by penalizing action and joint-limit violations to promote smooth and physically stable control. To improve training stability and efficiency, we employ early termination to stop obviously failing rollouts, and adaptive sampling~\cite{liao2025beyondmimic} to balance the learning of motion segments at each difficulty depending on their fail counts. In the following sections, we detail the differences between the two stages on the desired goal and learning configuration modifications based on these general settings. We provide full details on the general training setting in~\cref{sec:appendix:rl_training_details}.

\subsection{Stage I: Physical Motion Generation} \label{method:pmg} 

The first stage acts as a motion filtering process that generates \textbf{physically feasible reference motions} from raw retargeted data. Specifically, we train a single generalist policy in simulation to generate physically refined motion while maintaining the style of the original reference motion. To produce high-quality rollout trajectories for future policy learning, we provide the policy with both robot proprioceptive states and full privileged information, including global poses, orientations, velocities, contact states, and other simulation-only signals unavailable on real hardware. 
We use a relatively large early-termination threshold in this stage. Since raw references may contain physical artifacts, overly strict termination would incorrectly treat many near-reference states as infeasible and prevent sufficient coverage of motion segments during training.
And the availability of richer privileged observations improves motion realizability, allowing for a relatively higher sampling floor while maintaining a stable overall training distribution. We do not apply domain randomization in the physical motion generation stage. Instead, we inject noise into motion commands to encourage robust and stable rollouts across reference motions from diverse sources. The robot’s joint torque limits in this stage are still constrained by real hardware capabilities and remain consistent with the torque limits used in the subsequent control stage.
The rollout robot states from this stage are recorded as physically feasible reference trajectories, which serve as training targets for the next stage. We store not only the robot’s joint states but also global poses, orientations, velocities, and contact states as part of the physically consistent motion commands.

\subsection{Stage II: General Motion Tracking}  
\label{method:gmt}

The second stage addresses the challenge of \textbf{general motion tracking}.  Here, we train a general control policy using the physically feasible trajectories generated in Stage I. Since these references are already dynamics-consistent, the intrinsic conflict between tracking fidelity and physical stability is removed, and the learning problem reduces to motion tracking under uncertainty.
In contrast to Stage I, this stage is designed for real-world deployment. The robot state $\selfstate$ is restricted to proprioceptive inputs $(\simp,\simv,\bs{R}_{t},\simav,\bs{a}_{t-1})$ including joint pose $\simp$, joint velocity $\simv$, root orientation $\bs{R}_{t}$, root angular velocity $\simav$ and previous action $\bs{a}_{t-1}$, all of which are available in real world. 

Because the reference trajectories are generated from simulator rollouts and contain consistent contact information, we incorporate desired-contact rewards to encourage coherent and physically meaningful contact behaviors. We scale the action smoothness penalty with the reference joint velocities, permitting faster control changes in dynamic phases while enforcing stronger smoothing in slow or quasi-static motions, improving control smoothness and physical stability without sacrificing tracking performance. Detailed reward definitions are provided in \cref{tab:appendix:reward}.
A smaller minimum sampling weight is used in this stage. With reduced observation information compared to Stage I, certain motions become more difficult to track. Lowering the sampling floor ensures that such motions continue to receive sufficient training exposure despite their lower initial failure counts.
To reflect deployment conditions, we introduce observation noise and extensive domain randomization to improve sim-to-real robustness. Specifically, we randomize physical parameters such as friction, restitution, joint properties, IMU offsets, and center of mass, and apply random external pushes during training.

\subsection{Tasks and Applications} \label{method:application}
The two stages of \modelname form a flexible and integrated pipeline. While each stage can be used independently for different purposes, they can also be composed into a unified system that supports online inference, enabling applications such as real-time teleoperation.

\textbf{Offline motion filtering and tracking.}
In the offline setting, the physical motion generation stage is applied to one or multiple motion clips to produce physically feasible trajectories via simulator rollouts. This process serves as a motion filtering step that removes embodiment-induced artifacts and enforces dynamics consistency. The resulting general motion tracking policy can then be deployed as a high-quality motion tracker, capable of sustaining long-horizon, stable execution across diverse behaviors, including highly dynamic motions such as flips and cartwheels.

\textbf{Online teleoperation and motion tracking.}
In the online setting, the two stages are combined to form an end-to-end inference pipeline. Motion commands from motion capture systems, VR headset or other real-time sources are first processed by the physical motion generation module in simulation, such as IsaacLab~\cite{mittal2025isaac} or MuJoCo~\cite{todorov2012mujoco}, to produce physically valid and dynamically consistent trajectories. These refined motion commands are then fed into the general motion tracking policy, which outputs joint-level actions for the real robot in real time. This unified pipeline enables human-style, high-dynamic teleoperation as well as online motion tracking under continuously varying and potentially unpredictable user input.





\section{Experiment}
In this section, we first analyze the quality of the reference motions generated in simulation and their impact on policy learning~\cref{exp:physical_motion}. We then evaluate the general motion tracking policy in terms of motion coverage, tracking accuracy, and generalization in simulation~\cref{exp:gmt_sim}. Finally, we validate \modelname's robustness on real humanoid hardware across diverse dynamic behaviors and deployment scenarios~\cref{exp:gmt_real}.

\subsection{Quality of Generated Physically Plausible Motions.}
\label{exp:physical_motion}

We process two categories of source motion datasets to obtain physically plausible reference motions used throughout our experiments:

\begin{itemize}
\item Unitree-retargeted LAFAN1~\cite{harvey2020robust}, consisting of 40 motion clips, with a total duration of approximately 2.5 hours.
\item AMASS Subset~\cite{mahmood2019amass}, comprising approximately 10,000 motion capture sequences retargeted using GMR~\cite{araujo2025retargeting}.
\end{itemize}

These datasets provide a diverse set of human motions covering daily activities, locomotion, athletic movements, and complex whole-body behaviors. We apply our physical motion generation stage to all sequences in both datasets to obtain dynamically consistent reference motions.

\subsubsection{\textbf{Physically plausible motions with style fidelity}} 
We first evaluate the physical feasibility of the generated motions. As shown in \cref{fig:phsical artifacts} and \cref{tab:physical_motion}, motions produced by the physical motion generator eliminate common artifacts in raw retargeted data, such as penetration and floating, resulting in trajectories that are dynamically stable and consistent with the robot’s actuation limits. The generated motions also exhibit improved state smoothness. Although enforcing dynamics-consistent corrections results in a moderate increase in MPJPE, this increase reflects necessary adjustments toward physically realizable poses rather than degradation in motion quality. Despite these corrections, the generated motions retain the stylistic characteristics and temporal structure of the original motions, indicating a balance between physical feasibility and style fidelity. Metric definitions are provided in \cref{appendix:experiment_setting}

\begin{table}[!t]
    \caption{\textbf{Comparison of raw and physically plausible motions.} Physically plausible motions eliminate penetration and floating artifacts while improving motion smoothness, with a moderate increase in MPJPE due to dynamics-consistent corrections.}
    \label{tab:physical_motion}
    \centering
    \renewcommand{\arraystretch}{1.20}
    \setlength{\tabcolsep}{4.5pt}

    \resizebox{\columnwidth}{!}{
    \begin{tabular}{lccccc}
        \toprule
        \textbf{Dataset} & \textbf{Reference} 
        & \textbf{Penetration} 
        & \textbf{Floating} 
        & \textbf{Smoothness} 
        & \textbf{MPJPE} \\
        \midrule

        \multirow{2}{*}{{\small LAFAN1}}
            & Raw Ref.      
                & 20.3\%  & 2.52\% & 33.7 & 0.000 \\
            & Physical Ref. 
                & \textbf{0.0\%} & \textbf{0.0\%} & \textbf{31.8} & 21.0 \\
        \midrule

        \multirow{2}{*}{{\small AMASS}}
            & Raw Ref.      
                & 67.7\% & 3.29\% & 19.5 & 0.000 \\
            & Physical Ref. 
                & \textbf{0.0\%} & \textbf{0.0\%} & \textbf{15.8} & 16.0 \\
        \bottomrule
    \end{tabular}
    }
\end{table}

\begin{figure*}[!t]
    \centering
    \includegraphics[width=\textwidth]{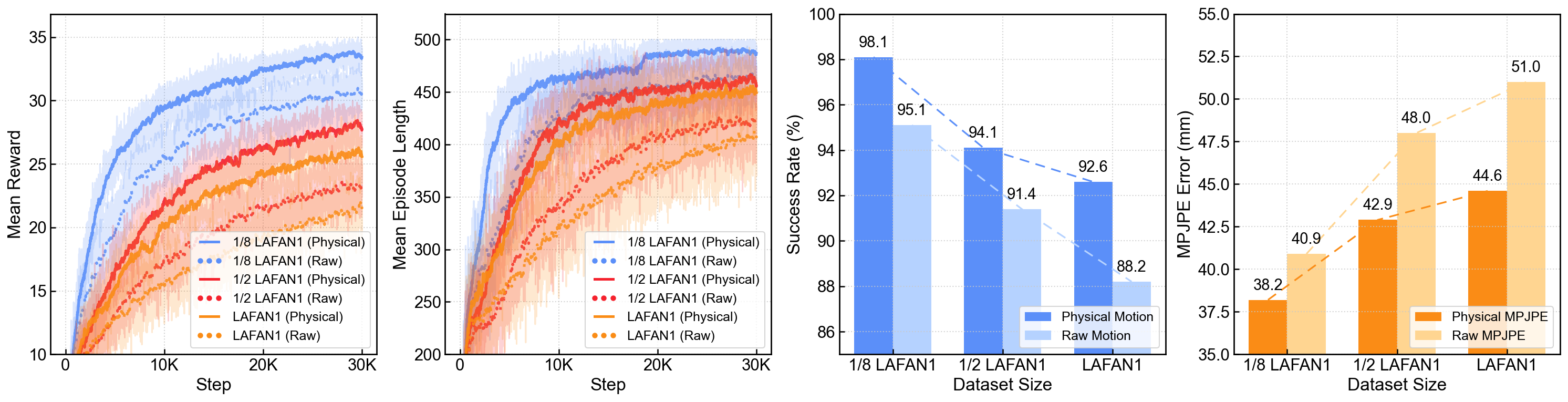}
    \caption{\textbf{Impact of physically plausible motions under varying dataset sizes.} From left to right: mean reward and mean episode length during training, followed by success rate and tracking error (MPJPE) under different dataset sizes (1/8, 1/2, and full LAFAN1). Dark colors denote training with physically plausible motions, while light colors denote training with raw reference motions.}
    \label{fig:curve}
\end{figure*}

\begin{table*}[!t]
\caption{\textbf{Comparison of training pipelines for motion tracking.} Our training pipeline consistently outperforms alternative training strategies across all motion categories, with particularly significant advantages on challenging motion sequences.}
\label{tab:pipeline_comparison}
\centering
\renewcommand{\arraystretch}{1.2}
\begin{tabular*}{\textwidth}{@{\extracolsep{\fill}}lcccccccccccc}
    \toprule
    \multirow{3}{*}{\textbf{Pipeline}} &
    \multicolumn{4}{c}{\textbf{Seen Motions}} &
    \multicolumn{4}{c}{\textbf{Seen (Hard Motions)}} &
    \multicolumn{4}{c}{\textbf{Unseen Motions}} \\
    \cmidrule(lr){2-5} \cmidrule(lr){6-9} \cmidrule(lr){10-13}
    & \textbf{SR}
    & \textbf{MPJPE}
    & $\boldsymbol{\Delta}$\textbf{Vel}
    & $\boldsymbol{\Delta}$\textbf{Acc}
    & \textbf{SR}
    & \textbf{MPJPE}
    & $\boldsymbol{\Delta}$\textbf{Vel}
    & $\boldsymbol{\Delta}$\textbf{Acc}
    & \textbf{SR}
    & \textbf{MPJPE}
    & $\boldsymbol{\Delta}$\textbf{Vel}
    & $\boldsymbol{\Delta}$\textbf{Acc} \\
    \midrule

    Dagger\cite{ross2011reduction}
    & 92.46\% & 40.79 & 6.196 & 2.051
    & 67.40\% & 53.52 & 9.468 & 3.152
    & 96.56\% & 37.34 & 5.113 & 1.598 \\

    Dagger$_{\text{hist}}$
    & 93.03\% & 38.96 & 5.714 & 1.945
    & 70.04\% & 52.15 & 8.808 & 3.027
    & 96.52\% & 35.66 & 4.863 & 1.559 \\

    AAC\cite{pinto2017asymmetric}
    & 92.78\% & 41.36 & 6.577 & 2.036
    & 68.80\% & 55.46 & 10.16 & 3.157
    & 96.55\% & 37.24 & 5.309 & 1.574 \\

    \textbf{Ours}
    & \textbf{96.13\%} & \textbf{37.61} & \textbf{5.709} & \textbf{1.934}
    & \textbf{84.81\%} & \textbf{46.43} & \textbf{8.516} & \textbf{2.932}
    & \textbf{96.88\%} & \textbf{34.83} & \textbf{4.852} & \textbf{1.556} \\

    \bottomrule
\end{tabular*}
    \vspace{-1em}
\end{table*}

\subsubsection{\textbf{Physically consistent reference motions enable stable and scalable policy learning}}\label{exp:physical_motion:physical_motion_scalable}

The physical consistency of reference motions directly affects the learning efficiency and final performance of motion tracking policies. To evaluate how this effect scales with dataset size, we conduct controlled experiments on the LAFAN1~\cite{harvey2020robust} dataset, comparing training with raw retargeted motions and with physically plausible motions generated by our method. All models are trained for 30K iterations, and scalability is evaluated by training on one-eighth, one-half and the full dataset. We further evaluate the trained policies in simulation under random external pushed applied to the robot's pelvis and report the success rate and MPJPE with respect to the original raw reference motions over 30K evaluation episodes in \cref{fig:curve} and \cref{tab:motion_pair_comparison}.

Training on raw retargeted motions leads to significant degradation in learning performance as dataset scale increases. Convergence becomes slower, final rewards decrease, and training stability worsens, with sharper declines in success rates and more severe deterioration in motion fidelity as motion diversity grows. For example, when scaling from one-eighth to the full dataset, the success rate drops from 95.1\% to 88.2\% and MPJPE increases by 10.1 mm.
In contrast, when trained on physically plausible motions, all reference trajectories remain dynamically feasible regardless of dataset size. Although increased motion diversity still raises task complexity, convergence slows only moderately and final rewards remain high. Success rates remain stable at around 92.6\%, and motion fidelity to the original references is largely preserved even when training on the full dataset, with MPJPE increasing by only 6.4 mm.

These results show that enforcing physical consistency at the reference level is critical for scalable humanoid motion learning, enabling stable policy training even on large and diverse motion datasets.

\subsection{General Motion Tracking Performance in Simulation}
\label{exp:gmt_sim}

\subsubsection{\textbf{Experiment setting}}
\label{exp:gmt_sim:setting}
We train a general motion tracking policy on the full LAFAN1~\cite{harvey2020robust} dataset together with approximately eight hours of CMU motions from AMASS~\cite{mahmood2019amass}, covering a broad spectrum of human motor skills. For generalization evaluation, we use approximately sixteen hours of motion data from the KIT subset as a test set. To further assess performance on highly dynamic behaviors, we define a challenging subset of 24 motion episodes within LAFAN1 by excluding simple locomotion sequences such as walking and running, retaining motions with richer dynamics and contact interactions. All methods, including ours and the baselines, operate under the same partial-observation input as defined in ~\cref{method:gmt} and use single-frame observations as input. For fair comparison, evaluation metrics (metric definitions are provided in~\cref{appendix:experiment_setting}) are computed with respect to the original reference motions.

\begin{figure*}[!t]
    \centering
    \includegraphics[width=\textwidth]{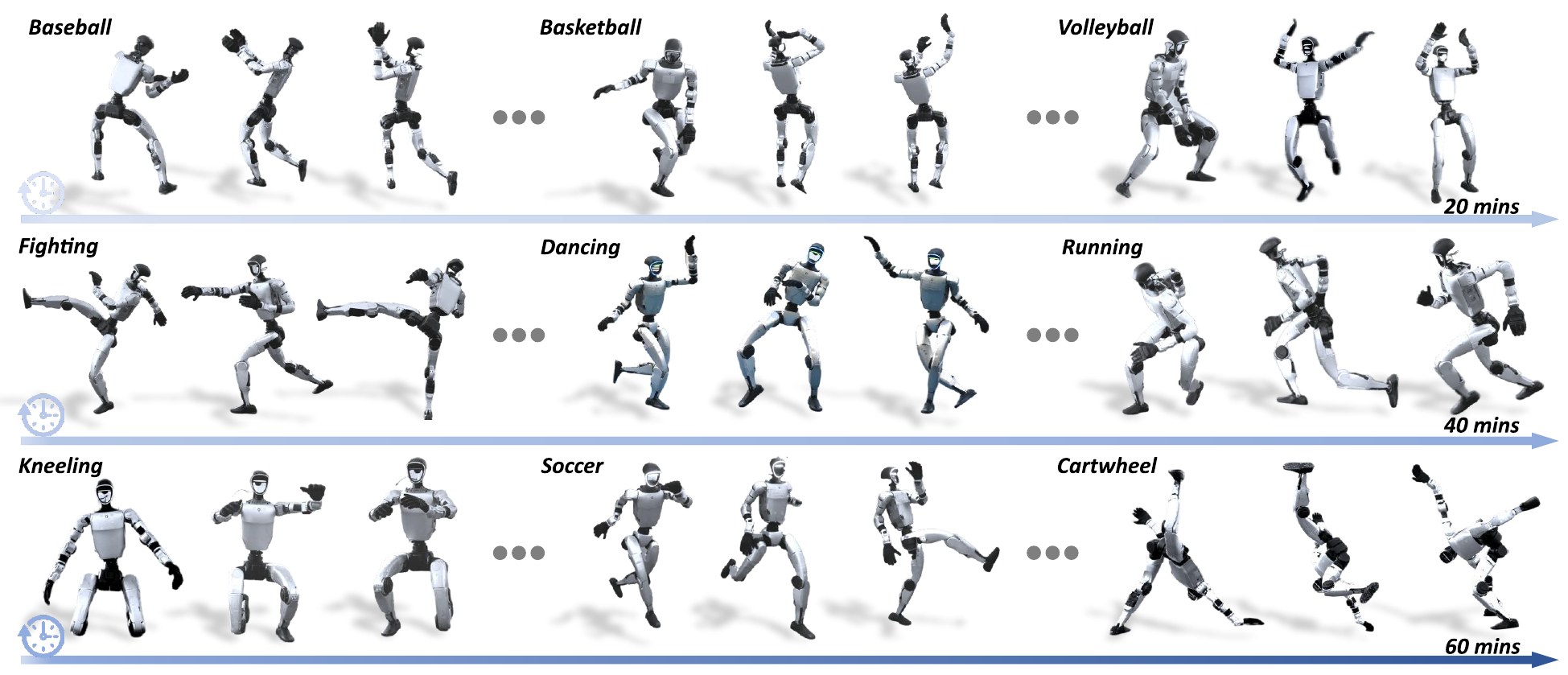}
    \caption{\textbf{Diverse motion skills executed on the real humanoid robot.} Our policy enables hour-long continuous and stable tracking of a wide range of human-like behaviors, demonstrating broad motion coverage, strong real-world versatility, and long-term control stability.}
    \label{fig:diverse_motions}
    \vspace{-1em}
\end{figure*}

\subsubsection{\textbf{Comparison of Training Pipelines}}
\label{:gmt_sim:pipeline}
We first compare different training pipelines to evaluate the impact of physically plausible reference motions. The baselines include: (1) a teacher–student\cite{ross2011reduction} framework (Dagger) that distills privileged information into a student policy, (2) an enhanced teacher–student variant (Dagger$_{\text{hist}}$) that incorporates a history window of length five, and (3) an asymmetric actor–critic\cite{pinto2017asymmetric} framework (AAC). All baseline policies are trained using the same raw motion inputs.

As shown in \cref{tab:pipeline_comparison}, our pipeline consistently outperforms all baselines across all evaluation metrics. On the seen motions, our method achieves a success rate of 96.13\% with an MPJPE of 37.61 mm, compared to the approximately 93\% success rate of the competing baselines. Improvements are even more pronounced on the high-dynamic subset, where baseline success rates drop sharply, while our method maintains stable success rate at 84.81\%. On the full test set, On the full test set, our method achieves the highest success rate of 96.88\% and the lowest MPJPE of 34.83 mm among all compared methods.

\textit{Structural decoupling leads to superior tracking performance.} 
Training on physically feasible reference trajectories allows the policy to focus exclusively on motion tracking rather than compensating for infeasible targets. In contrast, baselines trained on raw motions must implicitly learn to correct physical inconsistencies, which leads to unstable learning and degraded performance, especially in high-dynamic and contact-rich scenarios.


\subsubsection{\textbf{Comparison with State-of-the-Art Tracking Methods}}
\label{exp:gmt_sim:compare}
We further compare our method against representative state-of-the-art tracking approaches, including OmniH2O~\cite{he2024omnih2o}, ExBody2~\cite{ji2024exbody2}, and BeyondMimic~\cite{liao2025beyondmimic}. All approaches are re-implemented and trained in the IsaacLab~\cite{mittal2025isaac} simulator using identical training data to ensure a fair comparison. We report the performance of all methods on the challenging high-dynamic subset of LAFAN1, shown in \cref{tab:method_compare}. 
Our method achieves the best performance across all evaluation metrics. In particular, our policy attains a success rate of 84.81\%, outperforming OmniH2O (48.32\%), ExBody2 (58.93\%), and BeyondMimic (70.04\%). We also observe consistent reductions in MPJPE, along with improvements in velocity and acceleration tracking errors.

\textit{Physical inconsistency amplifies failure in high-dynamic motions.}
The performance gap becomes more pronounced in highly dynamic and contact-rich motions. Such behaviors impose strict dynamic constraints, where even small inconsistencies in reference trajectories can lead to compounding control errors. Baseline methods frequently exhibit unstable contact transitions and torque saturation, resulting in early failures. In contrast, our framework enforces dynamic validity at the reference level, preventing cascading errors and enabling stable tracking even under high momentum and rapid contact changes.

\begin{table}[!t]
    \caption{\textbf{Comparison with state-of-the-art humanoid motion tracking methods.} Our method achieves the best overall performance, showing higher success rates and lower tracking errors compared with prior approaches.}
    \label{tab:method_compare}
    \centering
    \renewcommand{\arraystretch}{1.20}
    \setlength{\tabcolsep}{12pt} 
    \resizebox{\linewidth}{!}{
        \begin{tabular}{lcccc}
            \toprule
            \textbf{Method} & \textbf{SR} & \textbf{MPJPE} & $\boldsymbol{\Delta}$\textbf{Vel} & $\boldsymbol{\Delta}$\textbf{Acc} \\
            \midrule
            
            OmniH2O\cite{he2024omnih2o} & 48.32\% & 54.58 & 11.40 & 4.293 \\
            Exbody2\cite{ji2024exbody2} & 58.93\% & 53.47 & 12.74 & 3.996 \\
            BeyondMimic\cite{liao2025beyondmimic} & 70.04\% & 55.75 & 10.04 & 3.126 \\
            \textbf{Ours} & \textbf{84.81\%} & \textbf{46.43} & \textbf{8.516} & \textbf{2.932} \\
            \bottomrule
        \end{tabular}
    }
\end{table}

\begin{figure*}[!t]
    \centering
    \includegraphics[width=\textwidth]{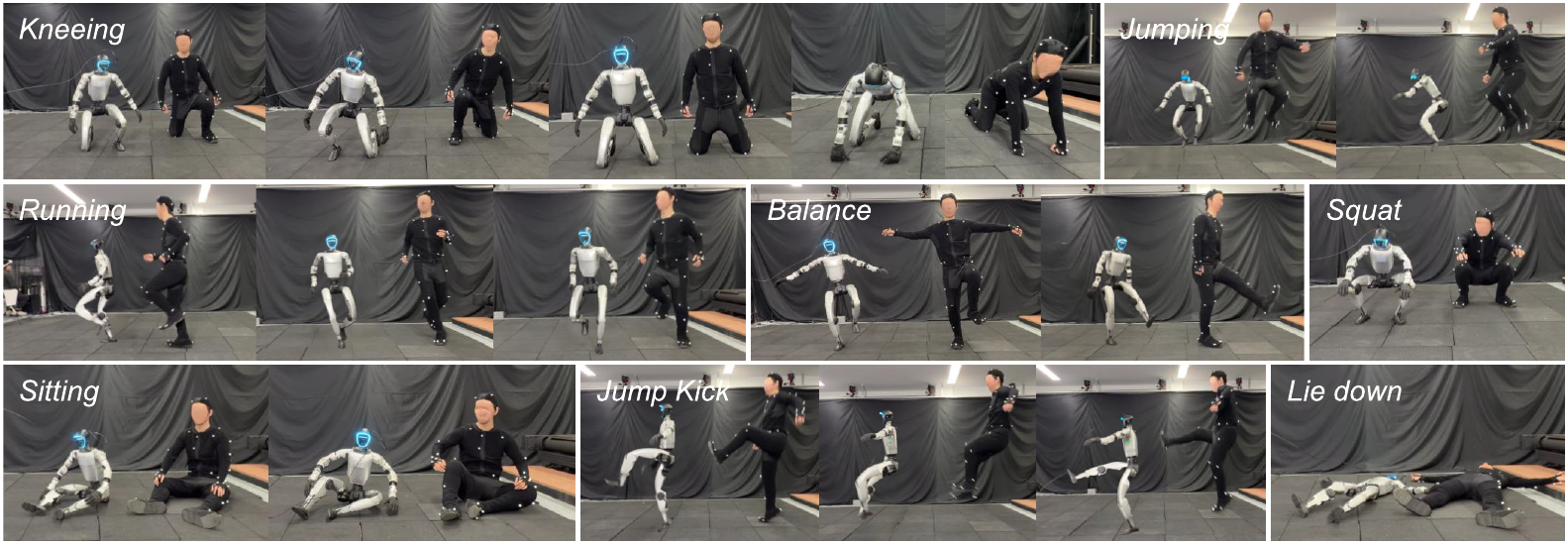}
    \caption{\textbf{Real-time teleoperation on the humanoid robot.} Under real-time teleoperation, the robot executes dynamic, static, and contact-rich behaviors with natural, human-like motion style, demonstrating responsive control and high-fidelity whole-body coordination in the real world.}
    \label{fig:teleopretion}
\end{figure*}

\subsection{Performance of General Motion Tracking in Real}
\label{exp:gmt_real}

\subsubsection{\textbf{Motion Tracking}}
\label{exp:gmt_real:motion_tracking}

To evaluate sim-to-real transfer, we assess the policy’s robustness, stability, and motion coverage on the physical robot.

\textbf{Diverse Real-World Skill Repertoire.} We deploy representative motion clips from the training set on the real robot and show that it can stably and continuously execute these diverse motions for over an hour, covering locomotion, jumping, dancing, martial arts, fall recovery, balance control, and highly dynamic maneuvers such as flips, as shown in \cref{fig:teaser,fig:diverse_motions}. We further provide quantitative evaluations of both sim-to-sim and sim-to-real tracking performance, as shown in~\cref{tab:mujoco_real}. We select six diverse motion clips and compare tracking errors in joint position, base orientation, and base angular velocity between MuJoCo~\cite{todorov2012mujoco} simulation and real-robot execution. To ensure a fair comparison, observation noise is injected into the MuJoCo measurements to match real-world sensor noise levels. The results show comparable tracking error levels across simulation and real-world deployments for all tested motions, indicating effective sim-to-real alignment.

\begin{table}[!t]
    \caption{\textbf{Sim-to-real tracking error comparison using onboard sensor measurements.} We evaluate motion tracking errors in both MuJoCo~\cite{todorov2012mujoco} simulation and the real robot using directly accessible and reliable sensor signals with noise modeling. The results show comparable error levels across domains for diverse motions, indicating effective sim-to-real alignment. Detailed definitions of the sensor signals are provided in~\cref{sec:appendix:rl_training_details}.}
    
    
    \label{tab:mujoco_real}
    \centering
    \renewcommand{\arraystretch}{1.15}

    \resizebox{\columnwidth}{!}{
    \begin{tabular}{lcccccc}
        \hline
        \multirow{2}{*}{\textbf{Motion}} &
        \multicolumn{3}{c}{\textbf{MuJoCo (Sim)}} &
        \multicolumn{3}{c}{\textbf{Real World}} \\
        \cline{2-7}
        & \textbf{Dof} & \textbf{Rot} & \textbf{AngVel} &
          \textbf{Dof} & \textbf{Rot} & \textbf{AngVel} \\
        & (rad) & (rad) & (rad/s) &
          (rad) & (rad) & (rad/s) \\
        \hline

        Walking & 0.0601 & 0.0675 & 0.6567 &
                   0.0710 & 0.0996 & 0.6716 \\

        Running & 0.0580 & 0.0730 & 0.6781 &
                  0.0638 & 0.0897 & 0.7304 \\


        Dancing & 0.0657 & 0.0752 & 0.7370 &
                  0.0708 & 0.0925 & 0.7613 \\

        Martial Arts & 0.0817 & 0.0928 & 0.7794 &
                      0.0822 & 0.0956 & 0.6730 \\
                      
        Getting up & 0.0753 & 0.1007 & 0.7097 &
                      0.0862 & 0.1410 & 0.8641 \\

        Cartwheel & 0.0821 & 0.0989 & 0.8905 &
                0.0937 & 0.1274 & 0.9316 \\

        \hline
    \end{tabular}
    }
\end{table}

\textbf{Robustness in Long-Horizon and Challenging Real-World Scenarios.} 
We further evaluate robustness under long-horizon execution and challenging outdoor environments, including uneven terrain, soft soil, and surfaces with varying friction. In these conditions, the robot continuously replays motion sequences outdoors for approximately 30 minutes while maintaining stable tracking under strong wind and external disturbances such as kicks and impacts.
To stress-test dynamic consistency, the robot performs 17 consecutive side flips without failure, demonstrating sustained control over extended high-dynamic sequences. Additional demonstrations are provided in the supplementary video.

\subsubsection{\textbf{Online Teleopretion}}
\label{exp:gmt_real:Teleopretion}

We evaluate real-time teleoperation to assess the policy’s generalization to unseen and unstructured motion commands.

\textbf{Robustness to Noisy Teleoperation Inputs.}
Human operators provide motion commands through two different teleoperation interfaces: a full-body motion capture system and a Pico VR headset~\cite{zhao2025xrobotoolkit}. The captured human motions are retargeted to the humanoid using the GMR retargetor~\cite{araujo2025retargeting}. These teleoperation inputs inevitably introduce noise, jitter, latency, and other artifacts, resulting in reference motions that may contain discontinuities and dynamically inconsistent segments. Without any manual preprocessing, the physical motion generation module converts these teleoperation commands into physically feasible trajectories, which are then tracked by the general motion tracking policy on hardware. As shown in \cref{fig:teleopretion}, the robot produces smooth, human-like, and dynamically stable motions while preserving the operator’s intended motion style. This remains true even with highly dynamic, high-amplitude and contact-rich teleoperation inputs. These results demonstrate strong robustness and generalization of our system to noisy, artifact-corrupted, and previously unseen real-world motion commands from heterogeneous teleoperation devices.

\section{Conclusion} 
\label{sec:conclusion}

We presented \modelname, a general humanoid motion tracking framework that resolves the conflict between tracking fidelity and physical feasibility caused by embodiment gaps and noisy human motion data. Our key idea is to decouple physical feasibility from general control learning via a two-stage pipeline: first generating dynamics-consistent reference motions, then training a general policy under realistic partial observability. This design eliminates the need for the controller to implicitly handle infeasible targets during execution. Extensive simulation and real-world experiments show improved tracking accuracy, strong generalization to unseen motions, and robust sim-to-real transfer. On a Unitree G1 humanoid robot, our method enables long-horizon stable deployment, highly dynamic skills, and responsive online teleoperation. These results demonstrate that explicitly resolving physical feasibility before policy learning is crucial for scalable and general humanoid control.



\clearpage
\bibliographystyle{plainnat}
\bibliography{references}

\clearpage

\appendix \label{sec:appendix}

\renewcommand{\thefigure}{A.\arabic{figure}}
\renewcommand{\thetable}{A.\arabic{table}}
\renewcommand{\theequation}{A.\arabic{equation}}

\subsection{Details on Policy Learning with Reinforcement Learning} \label{sec:appendix:rl_training_details}

We detail the reward design, domain randomization, adaptive sampling settings below.

\begin{table}[h]
\centering
\caption{General reward terms.}
\label{tab:appendix:reward}
\resizebox{\linewidth}{!}{
    \begin{tabular}{llc}
    \toprule
    \textbf{Reward Terms} & \textbf{Equation} & \textbf{Weight} \\
    \midrule
    \multicolumn{3}{l}{\emph{Tracking}}\\
    Body pos
    & $\displaystyle
    \exp\!\Big(
    -\big( \tfrac{1}{|\mathcal{B}|}
    \sum_{b\in\mathcal{B}}
    \|\mathbf{p}^{\mathrm{ref}}_{b}-\mathbf{p}_{b}\|^{2} \big) / 0.3^{2}
    \Big)$
    & $1.0$\\[2mm]
    Body ori
    & $\displaystyle
    \exp\!\Big(
    -\big( \tfrac{1}{|\mathcal{B}|}
    \sum_{b\in\mathcal{B}}
    \|\log(\phi^{\mathrm{ref}}_{b} \phi_{b}^{\top})\|^{2} \big) / 0.4^{2}
    \Big)$
    & $1.0$\\[2mm]
    Body lin. vel
    & $\displaystyle
    \exp\!\Big(
    -\big( \tfrac{1}{|\mathcal{B}|}
    \sum_{b\in\mathcal{B}}
    \|\mathbf{v}^{\mathrm{ref}}_{b}-\mathbf{v}_{b}\|^{2} \big) / 1.0^{2}
    \Big)$
    & $1.0$\\[2mm]
    Body ang. vel
    & $\displaystyle
    \exp\!\Big(
    -\big( \tfrac{1}{|\mathcal{B}|}
    \sum_{b\in\mathcal{B}}
    \|\boldsymbol{\omega}^{\mathrm{ref}}_{b}-\boldsymbol{\omega}_{b}\|^{2} \big) / 3.14^{2}
    \Big)$
    & $1.0$\\[2mm]
    Root pos (PMG)
    & $\displaystyle
    \exp\!\Big(
    -\|\mathbf{p}^{\mathrm{ref}}_{\text{root}}-\mathbf{p}_{\text{root}}\|^{2} / 0.3^{2}
    \Big)$
    & $0.5$\\[2mm]
    Root ori
    & $\displaystyle
    \exp\!\Big(
    -\|\log(\phi^{\mathrm{ref}}_{\text{root}}\phi_{\text{root}}^{\top})\|^{2} / 0.4^{2}
    \Big)$
    & $0.5$\\[2mm]
    Desired contacts (GMT)
    & $\sum_{b\in\mathcal{B}_{\mathrm{contact}}}
    \mathbf{1}\!\left[\|f_{b}\|>1  \text{N}\right]$
    & $0.1$\\
    \midrule
    \multicolumn{3}{l}{\emph{Regularization}}\\
    Action smoothness
    & $\|\mathbf{a}_{t}-\mathbf{a}_{t-1}\|^{2}$
    & $-0.1$\\
    Joint position limit
    & $\sum_{j=1}^{N}\mathbf{1}\big[q_j>q_{j,max}|q_j<q_{j,min}]$
    & $-10.0$\\
    Undesired contacts (PMG)
    & $\sum_{b\notin\mathcal{B}_{\mathrm{ee}}}
    \mathbf{1}\!\left[\|f_{b}\|>1  \text{N}\right]$
    & $-0.1$\\
    \bottomrule
    \end{tabular}
}
\end{table}

\textbf{Reward terms.} We adopt a common tracking reward formulation $r_t = r_{\text{track}} + r_{\text{reg}}$,
where $r_{\text{track}}$ penalizes deviations in root and body-link poses, orientations, and velocities with respect to the reference motion, and $r_{\text{reg}}$ regularizes the learning process by penalizing excessive actions and joint-limit violations, thereby encouraging smooth and physically stable control.

The detailed reward terms are summarized in ~\cref{tab:appendix:reward}. Let $\mathcal{B}$ denote the set of tracked body links. During the physical motion generation stage, we set $\mathcal{B} = 14$, tracking the major body links, including the thighs, shanks, knees, torso, shoulders, and arms. In contrast, during the general motion tracking stage, we track all 31 body links of the G1 humanoid. This is feasible because all physical motions at this stage are obtained via rollout in simulation and thus inherently satisfy both physical feasibility and the kinematic constraints of the G1 humanoid.

In addition, we introduce a \textbf{desired contacts} reward during the general motion tracking stage. Since motion command rollouts are performed in simulation, we have access to accurate per-link contact forces and contact masks, which enables explicit supervision of contact behaviors.

\begin{table}[h]
\centering
\caption{Domain randomization parameters. ($\mathcal{U}[\cdot]$: uniform distribution)}
\label{tab:appendix:domain_rand}

\resizebox{\linewidth}{!}{
    \begin{tabular}{ll}
    \toprule
    \textbf{Domain Randomization} & \textbf{Sampling Distribution} \\
    \midrule
    \multicolumn{2}{l}{\emph{Physical parameters} (GMT)}\\
    Static friction coefficients & $\mu_{\text{static}} \sim \mathcal{U}[0.3,\,1.6]$ \\
    Dynamic friction coefficients & $\mu_{\text{dynamic}} \sim \mathcal{U}[0.3,\,1.2]$ \\
    Restitution coefficient & $e_{\text{rest}} \sim \mathcal{U}[0,\,0.5]$ \\
    Default joint pos [rad] & $ \Delta q^{default} \!\sim  \mathcal{U}[-0.01,\,0.01] $\\
    Default root ori [rad] & $ \Delta\phi_{r,roll} \!\sim  \mathcal{U}[-0.02,\,0.02]$, \\ & $ \Delta\phi_{r,pitch} \!\sim  \mathcal{U}[-0.02,\,0.02]$, \\ & $ \Delta\phi_{r,yaw} \!\sim  \mathcal{U}[-0.02,\,0.02]$, \\
    Torso's COM offset [m]& $\Delta x\!\sim\!\mathcal{U}[-0.025,0.025]$, \\ &$ \Delta y\!\sim\!\mathcal{U}[-0.05,0.05]$, \\ &$ \Delta z\!\sim\!\mathcal{U}[-0.05,0.05]$ \\
    \midrule
    \multicolumn{2}{l}{\emph{Root velocity perturbations} (GMT)}\\
    Root lin. vel [m/s]& $v_x, v_y\!\sim\!\mathcal{U}[-0.5,0.5]$, \\& $ v_z\!\sim\!\mathcal{U}[-0.2,0.2]$ \\
    Root ang. vel [rad/s]& $\omega_x,\ \omega_y\!\sim\!\mathcal{U}[-0.52,0.52]$, \\& $\omega_z\!\sim\!\mathcal{U}[-0.78,0.78]$ \\
    Push duration [s]& $\Delta t \sim \mathcal{U}[1.0,\,2.0]$ \\
    \midrule
    \multicolumn{2}{l}{\emph{Reference motion perturbations} (PMG)}\\
    Ref joint pos [rad]& $\Delta q^{ref}\!\sim\!\mathcal{U}[-0.01,0.01]$ \\
    Ref joint vel [rad/s]& $\Delta \dot{q}^{ref}\!\sim\!\mathcal{U}[-0.5,0.5]$ \\
    Ref root pos [m]& $\Delta x^{ref},\ \Delta y^{ref},\ \Delta z^{ref} \sim \mathcal{U}[-0.01,\,0.01]$ \\
    Ref root ori [rad]& $\Delta \phi^{ref}_{r,roll} \sim \mathcal{U}[-0.05,\,0.05]$, \\ &
    $\Delta \phi^{ref}_{r,pitch} \sim \mathcal{U}[-0.05,\,0.05]$, \\ & $\Delta \phi^{ref}_{r,yaw} \sim \mathcal{U}[-0.05,\,0.05]$ \\
    \bottomrule
    \end{tabular}
    }
\end{table}

\textbf{Domain randomization.} As shown in~\cref{tab:appendix:domain_rand}, we do not apply domain randomization during the physical motion generation stage. Instead, we inject noise into the motion commands, including the reference joint positions, reference joint velocities, reference root position, and reference root orientation, to encourage robust and stable rollouts across reference motions from diverse sources.

To better reflect real-world deployment conditions, we introduce observation noise and extensive domain randomization during the general motion tracking stage to improve sim-to-real robustness. Specifically, we randomize physical parameters such as friction coefficients, restitution, joint properties, IMU offsets, and the center of mass, and apply random external pushes during training.

\begin{table}[h]
\centering
\caption{Adaptive sampling parameters.}
\label{tab:appendix:domain_rand}

\resizebox{\linewidth}{!}{
    \begin{tabular}{ll}
    \toprule
    \textbf{Adaptive sampling} & \textbf{Value} \\
    \midrule
    Bin size & $1s$ \\
    Blending hyperparameter  & $\alpha=0.001$ \\
    Sampling prob clamp min/max   & $ [0.75/ N_{bin} , 100/ N_{bin}]$ \\
    \bottomrule
    \end{tabular}
    }
\end{table}

\textbf{Adaptive sampling.} We employ an adaptive sampling strategy to balance the learning of motion segments with different levels of difficulty, based on their failure statistics. Specifically, all motion sequences are concatenated and segmented into fixed-length bins of $1\,\mathrm{s}$. If the remaining tail of a motion sequence is shorter than $1\,\mathrm{s}$, it is still treated as an individual bin.

During training, we track the failure count of each bin and use these statistics to determine the sampling probability distribution. To avoid under-sampling easy bins or excessively over-sampling difficult ones, we clamp the sampling probability of each bin within a predefined range. Concretely, the lower and upper bounds are set to $0.75$ times and $100$ times the average sampling probability, respectively.

\begin{table*}[!t]
    \caption{Effect of Dataset Size and Reference Type on Controller Performance}
    \label{tab:motion_pair_comparison}
    \centering
    \renewcommand{\arraystretch}{1.3}
    \resizebox{\linewidth}{!}{
        \begin{tabular}{l c cccc c cccc c cccc}
            \toprule
            \multirow{2.5}{*}{Input Command / Ground Truth} & & \multicolumn{4}{c}{\textbf{5 Clips}} & & \multicolumn{4}{c}{\textbf{20 Clips}} & & \multicolumn{4}{c}{\textbf{40 Clips}} \\
            \cmidrule(lr){3-6} \cmidrule(lr){8-11} \cmidrule(lr){13-16}
             & & \small SR & \small MPJPE & \small $\Delta$vel & \small $\Delta$acc & & \small SR & \small MPJPE & \small $\Delta$vel & \small $\Delta$acc & & \small SR & \small MPJPE & \small $\Delta$vel & \small $\Delta$acc \\
            \midrule

            Raw reference / Raw reference & & 
            95.11\% & 40.93 & 7.445 & 2.460 & &
            91.45\% & 48.01 & 9.363 & 3.0292 & &
            88.18\% & 51.02 & 9.336 & 2.938 \\

            Physical reference / Raw reference & & 
            \textbf{98.11\%} & \textbf{38.23} & \textbf{6.412} & \textbf{2.360} & &
            \textbf{94.09\%} & \textbf{42.86} & \textbf{7.905} & \textbf{2.799} & &
            \textbf{92.57\%} & \textbf{44.64} & \textbf{7.880} & \textbf{2.709} \\
            
            \bottomrule
        \end{tabular}
    }
\end{table*}

\begin{figure*}[!t]
    \centering
    \includegraphics[width=\textwidth]{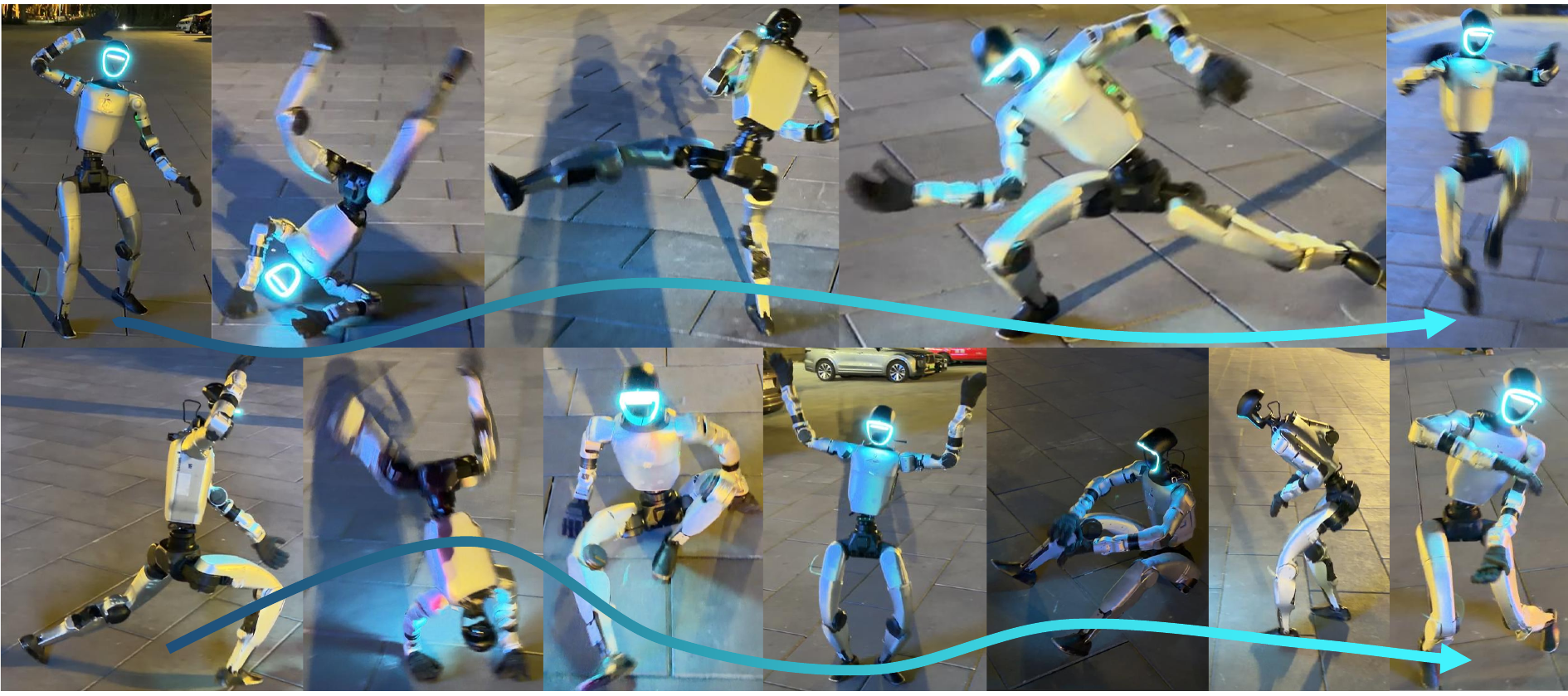}
    \caption{\textbf{One-hour continuous outdoor motion tracking without reset.} Starting from a fully charged battery, the robot continuously performs motion tracking in an outdoor environment for one hour until the battery is depleted. Please refer to the supplementary video demo for the full execution.
}
    \label{fig:teleopretion}
\end{figure*}

\begin{figure*}[!t]
    \centering
    \includegraphics[width=\textwidth]{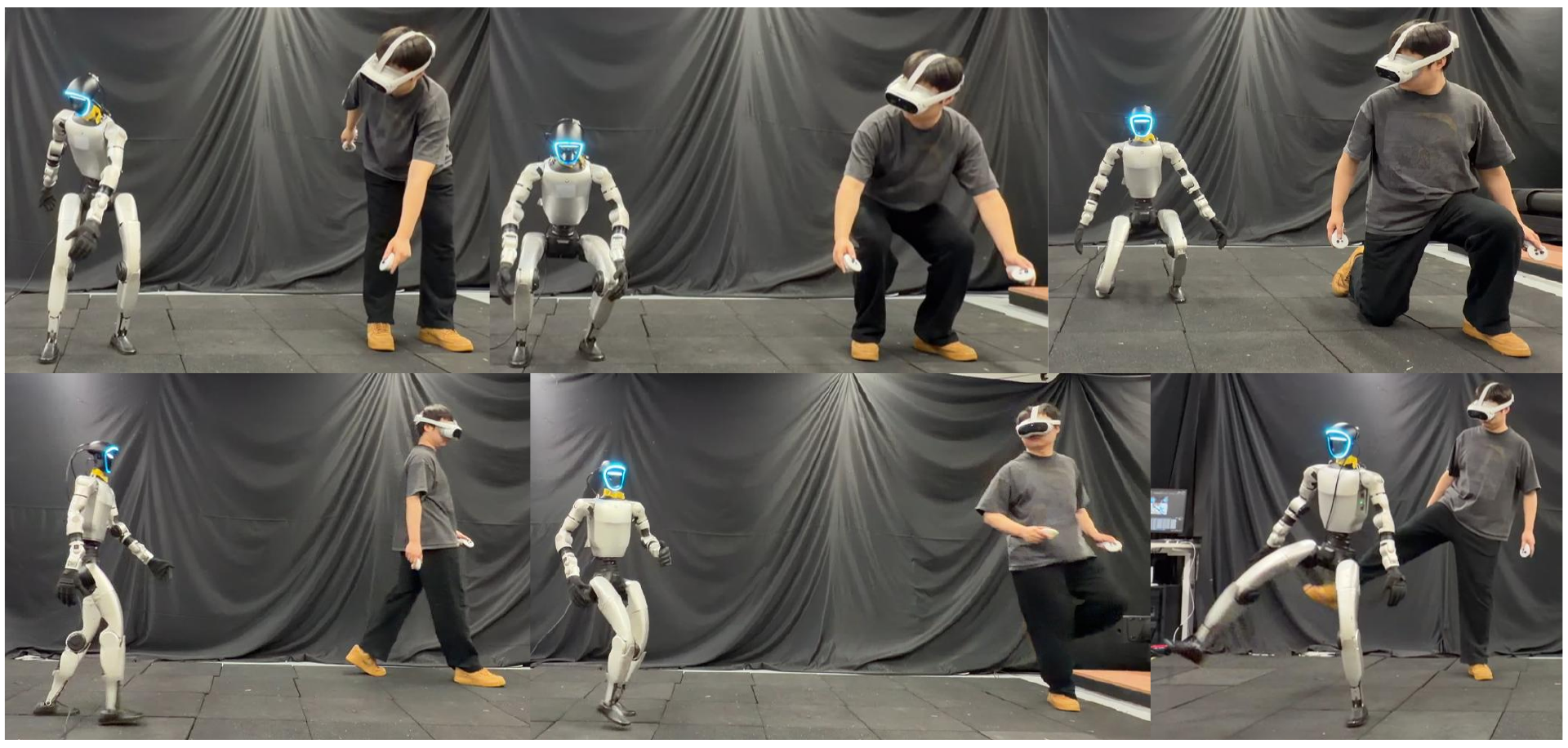}
    \caption{\textbf{VR Headset Real-time Teleoperation on the Humanoid Robot.} 
The raw SMPL motion is captured using the Pico VR Headset~\cite{zhao2025xrobotoolkit} and then retargeted to the humanoid robot via GMR~\cite{araujo2025retargeting}. The retargeted motion is further processed by the physical motion generation stage to produce physically consistent motions, which are finally tracked by the robot through the general motion tracking stage, enabling real-time teleoperation.}

    \label{fig:teleopretion}
\end{figure*}

\subsection{Experiment Setting}
\label{appendix:experiment_setting}

In all experiments, we use the Unitree G1 humanoid as the unified platform for both simulation and real-world deployment, leveraging all 29 degrees of freedom for control and evaluation. All policies are trained in IsaacLab~\cite{mittal2025isaac} with a 200 Hz physics stepping rate and a 50 Hz control frequency.

\textbf{Evaluation metrics of motion quality} The generated motions are evaluated with respect to both physical validity and fidelity to the original reference, using the following metrics:

\begin{itemize}
    \item \textbf{Penetration duration}: the cumulative time during which any part of the robot penetrates the environment or itself.
    \item \textbf{Floating duration}: We classify a motion segment as exhibiting a floating artifact when the root height remains above 0.8 m continuously for more than one second.
    \item \textbf{Trajectory smoothness}: defined as the linear jerk obtained by differentiating joint accelerations over time. This metric quantifies abrupt dynamical changes along the joint trajectories.
    \item \textbf{Style Fidelity}: measures how closely the physically generated motion matches the original reference motion. It is quantified using the mean per-joint position error, which evaluates the spatial differences between corresponding joint positions across the motion sequence.
\end{itemize}

\textbf{Evaluation metrics of Policy performance.} Policy performance is assessed using the following motion-tracking metrics:

\begin{itemize}
    \item \textbf{Success Rate}:  
    During evaluation, each motion sequence is re-segmented into 10-second clips.  
    A clip is marked as a failure if any of the following thresholds is exceeded:
    \begin{itemize}
        \item end-effector height error $>$ 0.25 m,
        \item root height error $>$ 0.25 m,
    \end{itemize}

    \item \textbf{Mean Per-joint Position Error (MPJPE, mm)};
    \item \textbf{Mean Per-joint linear velocity error ($\boldsymbol{\Delta}$Vel, mm/s)};
    \item \textbf{Mean Per-joint angular velocity error ($\boldsymbol{\Delta}$Acc, rad/s)}.
\end{itemize}

\subsection{Effect of Dataset Size and Reference Type on Controller Performance}
\cref{tab:motion_pair_comparison} reports additional quantitative results supporting the scalability analysis discussed in ~\cref{exp:physical_motion:physical_motion_scalable}. Specifically, we compare policies trained using raw retargeted motions and physically consistent reference motions generated by our method on the LAFAN1~\cite{harvey2020robust} dataset. The policies are evaluated in simulation under random external pushes applied to the robot’s pelvis, and performance is measured by the success rate and MPJPE with respect to the original raw reference motions, averaged over 30K evaluation episodes.

\end{document}